\documentclass[a4paper,11pt,twocolumn,twoside]{article}
\usepackage[dvips]{graphicx}
\usepackage{sepln_en}
\usepackage{fullname}
\usepackage[utf8]{inputenc} % allow utf-8 input
\usepackage[T1]{fontenc}    % use 8-bit T1 fonts
\PassOptionsToPackage{hyphens}{url}      % hyperlinks
\usepackage{url}            % simple URL typesetting
\usepackage{booktabs}       % professional-quality tables
\usepackage{amsfonts}       % blackboard math symbols
\usepackage{nicefrac}       % compact symbols for 1/2, etc.
\usepackage{microtype}      % microtypography
\usepackage{lipsum}
\usepackage{graphicx}
\graphicspath{ {./images/} }
\usepackage{booktabs}
\usepackage{float}
\usepackage{multirow}
\usepackage{color,soul}
\usepackage[inline]{enumitem}
\usepackage{tabularx} 
\usepackage{censor}
\usepackage{subfigure}

\newcommand{\anonym}[1]{#1}

\input epsf
\makeatletter
\newcommand{\printfnsymbol}[1]{%
  \textsuperscript{\@fnsymbol{#1}}%
}
\makeatother

\title{MarIA: Spanish Language Models}
\author{
  \textbf{Asier Gutiérrez-Fandiño\thanks{Equal contribution.},$^1$}
  \textbf{Jordi Armengol-Estapé\printfnsymbol{1},$^1$}
  \textbf{Marc Pàmies,$^1$}\\
  \textbf{Joan Llop-Palao,$^1$}
  \textbf{Joaquín Silveira-Ocampo,$^1$}
  \textbf{Casimiro Pio Carrino,$^1$}\\
  \textbf{Carme Armentano-Oller,$^1$}
  \textbf{Carlos Rodriguez-Penagos,$^1$}\\
  \textbf{Aitor Gonzalez-Agirre,$^1$}
  \textbf{Marta Villegas$^1$}\\
  $^1$Barcelona Supercomputing Center\\
  \texttt{marta.villegas@bsc.es} \\
}

\seplntranstitle{MarIA: Modelos del Lenguaje en Español}
\seplnclave{\anonym{MarIA}, Modelos de lenguaje del Español, Recursos de lenguaje del Español, Evaluación de modelos del lenguaje.}

\seplnresumen{
En este artículo se presenta \anonym{MarIA}, una familia de modelos del lenguaje en español y sus correspondientes recursos que se hacen públicos para la industria y la comunidad científica. Actualmente \anonym{MarIA} incluye los modelos del lenguaje en español RoBERTa-base, RoBERTa-large, GPT2 y GPT2-large que pueden considerarse como los modelos más grandes y mejores para español. Los modelos han sido preentrenados utilizando un corpus masivo de 570GB de textos limpios y deduplicados, que comprende un total de 135 mil millones de palabras extraidas del Archivo Web del Español construido por la Biblioteca Nacional de España entre los años 2009 y 2019. Evaluamos el rendimiento de los modelos con nueve conjuntos de datos existentes y con un nuevo conjunto de datos de pregunta-respuesta extractivo creado ex novo. El conjunto de modelos de \anonym{MarIA} supera, en la practica totalidad, el rendimiento de los modelos existentes en español en las diferentes tareas y configuraciones presentadas.
}

\seplnkey{\anonym{MarIA}, Spanish language modelling, Spanish language resources, Benchmarking.}
\seplnabstract{
This work presents \anonym{MarIA}, a family of Spanish language models and associated resources made available to the industry and the research community. Currently, \anonym{MarIA} includes RoBERTa-base, RoBERTa-large, GPT2 and GPT2-large Spanish language models, which can arguably be presented as the largest and most proficient language models in Spanish. The models were pretrained using a massive corpus of 570GB of clean and deduplicated texts with 135 billion words extracted from the Spanish Web Archive crawled by the National Library of Spain between 2009 and 2019. We assessed the performance of the models with nine existing evaluation datasets and with a novel extractive Question Answering dataset created ex novo. Overall, \anonym{MarIA} models outperform the existing Spanish models across a variety of NLU tasks and training settings.
}

\firstpageno{1}

\begin{document}

\label{firstpage} \maketitle
\setcounter{footnote}{0} 

\section{Introduction}

\label{sec:intro}
In recent years, the field of Natural Language Processing (NLP) has seen a proliferation of massive pretrained language models. These have been proved to perform best when trained on language-specific data. However, the vast majority of these massive models have been trained for English, leaving other languages aside and increasing the existing gap between them. Spanish, despite being the second most spoken language in the world, lacks large language models trained with vast and high quality data. One of the objectives of the \anonym{Plan-TL}\footnote{\url{https://plantl.mineco.gob.es/}}
is to cover this gap with the \anonym{MarIA} project.\footnote{\url{https://github.com/PlanTL-GOB-ES/lm-spanish}}
\anonym{MarIA} aims to provide both the industry and the scientific community with large scale language models, massive high-quality corpora and evaluation sets for the Spanish language. We present four large models of varying sizes and configurations, and compare them to existing models in a wide range of NLP tasks, showing that these new models are able to generalize better overall. 

The aim of this paper is to present an exhaustive report of all the work performed in the context of the \anonym{MarIA} project, which includes:
\begin{itemize}
    \item Processing of the largest \textit{clean} Spanish corpus to date, obtained from the web crawlings performed by the National Library of Spain from 2009 to 2019, used to 
    \item Train RoBERTa-base and RoBERTa-large models \cite{liu2019roberta}, and
    \item Train GPT2 and GPT2-large models \cite{radford2019language}.
    \item Creation of SQAC, a newly produced dataset for Spanish Question Answering.
    \item Conduction of a complete evaluation on a diverse set of tasks.
    \item Release of all  pre-trained and fine-tuned models in \url{https://huggingface.co/PlanTL-GOB-ES/}
\end{itemize}

The remainder of this paper is organized as follows. In Section \ref{sec:related_work}, we briefly go through the previous work done in language modeling, focusing on Spanish. In Section \ref{sec:data}, we describe the datasets used in the model training and in the subsequent evaluation. We devote special attention to the description of the training corpus and the new data set, expressly generated, on Question Answering. In Section \ref{sec:models} and \ref{sec:eval} we describe the new RoBERTa and GPT2 models and report in detail the evaluation methodology used and the eventual results. Finally, we present our conclusions and suggestions for future work in Section \ref{sec:conclusions}.
%The remainder of this paper is organized as follows. In Section \ref{sec:related_work}, we briefly go through the previous work done in Natural Language Processing (NLP) focusing on language modeling in the case of Spanish. In Section \ref{sec:data}, we describe the datasets we used. In Section \ref{sec:models}, we describe the new RoBERTa and GPT2 models. In Section \ref{sec:eval}, we evaluate the new models and compare their results with strong multilingual and monolingual baselines. Finally, we present our conclusions and suggestions for future work in Section \ref{sec:conclusions}.

%The remainder of this paper is organized as follows. In Section \ref{sec:related_work}, we analyze the previous work done in Natural Language Processing (NLP) focusing on the Spanish language. In Section \ref{sec:data}, we describe the new datasets used. In Section \ref{sec:models}, we describe the new RoBERTa models. In Section \ref{sec:eval}, we evaluate the new models and compare their results with strong multilingual (mBERT) and monolingual (BETO) models. In addition, we compare it to the BERTIN model\footnote{\url{https://huggingface.co/flax-community/bertin-roberta-large-spanish}} produced in the Flax/Jax Community Week.\footnote{\url{https://discuss.huggingface.co/t/open-to-the-community-community-week-using-jax-flax-for-nlp-cv/7104}} Finally, we present our conclusions and suggestions for future work, in Section \ref{sec:conclusions}.

\section{Related Work}
\label{sec:related_work}
Unsupervised pretraining started with the task of language modeling \cite{bengio2000neural}, where neural networks were trained to predict the next word from a given sequence, creating fixed vector representations known as word embeddings. Transfer learning capabilities of word embeddings took off with the introduction of Word2Vec \cite{mikolov2013efficient}, GloVe \cite{pennington-etal-2014-glove} and FastText \cite{bojanowski2016enriching}. For Spanish, researchers built datasets \cite{cardellinoSBWCE,banon-etal-2020-paracrawl,carrino2021spanish,jose_canete_2019_3247731} and computed word representations \cite{aitor_almeida_2018_1155474,doi:10.1177/1550147718811827,temu2021spanish,gutierrezfandino2021spanish} using those algorithms. %As words might contain typos, be made artificially, are not present on the training corpora (e.g. verbs in all the forms), or the text can contain words from other languages, word vectors started to be enriched with sub-word information \cite{10.1162/tacl_a_00051, heinzerling2018bpemb, DBLP:journals/corr/abs-1808-06226}. This is so relevant that all the Transformed models break words into smaller pieces to capture as much information as possible and to avoid unknown words.

Later on, researchers scaled up this unsupervised pretraining to larger datasets and more expressive models, specifically with language models, originally with LSTM-based \cite{10.1162/neco.1997.9.8.1735} models \cite{Peters:2018}. Nowadays, they are typically based on the Transformer architecture \cite{NIPS2017_3f5ee243}, with BERT \cite{DBLP:journals/corr/abs-1810-04805} as the paradigmatic example in the case of encoder models and the GPT family \cite{radford2018improving,Radford2019,gpt3} in the case of the decoder ones.

While the first models were either English-only or multilingual \cite{DBLP:journals/corr/abs-1810-04805}, researchers soon realized that building language-specific models was worth the effort \cite{camembert,flaubert,finnishbert,phobert,bertje,chinesebert}, provided there was enough data available. The language-specific literature with respect to language modeling has been quite prolific ever since \cite{DBLP:journals/corr/abs-2003-02912}. In the case of Spanish, the first BERT-based model was BETO \cite{jose_canete_2019_3247731}, which outperformed the strong multilingual baseline of mBERT.\footnote{The multiligual version of BERT.} BETO was trained on a collection of existing corpora, including the OPUS corpus \cite{opus} and the Spanish portion of Wikipedia. After the release of BETO, a few other models were published among which stands BERTIN\footnote{\url{https://huggingface.co/bertin-project/bertin-roberta-base-spanish/tree/v1-512}}, a series of Transformer-based models trained on the Spanish portion of the mC4 dataset \cite{mt5}.  %Recent research has shown that language models performance scales to very large numbers of parameters \cite{gpt3, shoeybi2020megatronlm, fedus2021switch}, provided there is enough data, which we intend to bring to the case of Spanish.

%Inspired by the previous architectures and believing that this direction hasn't more to offer, we developed both encoder and decoder models. As for encoders, we opted for the RoBERTa architecture \cite{liu2019roberta}, an improved version of BERT, and in the case of the decoders, we chose GPT2 \cite{Radford2019}.

Inspired by previous work carried out for different languages, we processed a new dataset and developed both new encoder and decoder models for Spanish. As for encoders, we opted for the RoBERTa architecture \cite{liu2019roberta}, an optimized version of BERT, and in the case of the decoders, we chose GPT2 \cite{Radford2019}. Further details are provided in the following sections.

\section{Data}
\label{sec:data}

This section describes the corpus used to pretrain the language models as well as the datasets used to evaluate them.

\subsection{Pretraining corpus}

The National Library of Spain (Biblioteca Nacional de España or BNE\footnote{\url{http://www.bne.es/en/Inicio/index.html}}) performs a crawling of all \texttt{.es} domains once a year. Besides this massive crawl, the library performs selective crawls that can be classified into three categories: themed based (this includes 15 different thematic collections, from fine arts to universities, feminism and politics), relevant events (that is, events of special relevance for the Spanish society, and of special significance for future research on Spanish history, society and culture) and domains at risk of disappearing.\footnote{\url{http://www.bne.es/en/Colecciones/ArchivoWeb/Subcolecciones/selectivas.html}}

We base our new pretraining corpus solely on these BNE's crawls carried out between the years 2009 and 2019. This means that sources that typically compose pretraining corpus of language models, such as Wikipedia, are not part of the dataset. This will have an effect on the evaluation, as we will see in Section \ref{sec:eval}.
% "compose" no es muy idiomático. Yo lo cambiaría por "feed"
Due to the massive amount of data, the National Library ran the first data extraction from WARC formatted files using the Selectolax Python library\footnote{\url{https://pypi.org/project/selectolax}} in its own premises. This process generated 59TB of JSON files containing some metadata along with the text extracted from the WARC files, namely: paragraphs, headers and hyperlinks' texts. 

To ensure the high quality of our training data, we developed an in-house cleaning pipeline inspired by the heuristics proposed in \cite{finnishbert}. It is composed of the following components:
\begin{enumerate}
    \item \textbf{Data parsing}: We parse text in different formats (e.g. CommonCrawl's WARC) keeping document-level boundaries.
    \item \textbf{Encoding detection and fixing}: We use \texttt{chardet}\footnote{\url{https://github.com/chardet/chardet}} to detect the encoding of the text and convert it to UTF-8 if required. Then, we apply \texttt{ftfy} \cite{speer-2019-ftfy}, a heuristic tool to fix common encoding errors.
    \item \textbf{Character document-level filtering}: We apply simple, inexpensive heuristics to discard lower quality documents. For example, we discard documents that are too short or those with too many characters associated to code snippets to prevent the inclusion of documents that are mainly Javascript snippets. We also apply a fast language identifier based on FastText \cite{bojanowski2016enriching}. Finally, we apply some regex-based rules to remove or transform placeholder text.
    \item \textbf{Sentence splitting}: We apply a heuristic sentence splitter.\footnote{\url{https://pypi.org/project/sentence-splitter/}} The heuristics are based on basic regex rules that account for acronyms (e.g., R.A.E. is not split in 3 different sentences).
    \item \textbf{Sentence-level filtering}: In this step, we apply more complex, fine-grained rules to discard some sentences within a document. The rationale is that in documents good-enough to get past the previous filters, there might be some sentences spoiling it, mainly coming from placeholder text or non-natural text. Thus, we execute a \textit{cascade} of language identifiers, that is, we first apply the fast (but less accurate) language identifier (FastText) with a relatively low confidence score, to minimize the number of false negatives (negative of being Spanish). Then we apply a slower but more accurate (in our preliminary tests) language identifier\footnote{\url{https://github.com/saffsd/langid.py}} to the sentences that passed the first language filter.
    \item \textbf{Deduplication}: We deduplicate text using Onion's \cite{pomikalek2011removing} N-gram-based deduplication. That is, for each document, Onion indexes 5-grams and marks as duplicates those documents whose overlapping in terms of 5-grams meets a certain threshold.
    \item \textbf{Formatting}: We write documents in plain text ensuring that document boundaries are kept.
\end{enumerate}
Note that we both transform and delete text. In the case of the encoding fixer, we apply transformations. In the case of the character-level document filter, we apply both transformations and deletions. In the case of sentence-level filter, language identification, and deduplication, we delete the text detected as low-quality, not Spanish, or duplicated.
% To ensure the quality of the data, we developed a cleaning pipeline which splits data into sentences, detects the language, removes noisy and ill-formed sentences (based on some heuristics \textbf{TODO: EXPLICAR}), deduplicates and eventually outputs the data with their original document boundaries. The pipeline is inspired by the heuristics proposed in \cite{finnishbert}.
The cleaning process took 96 hours in an HPC environment composed of 100 compute nodes, each with 48 CPU cores. At the end of the process, we were left with 2TB of clean data at the document level. Finally, after deduplication, we obtained a total of 570GB with more than 200M documents and 135B tokens of high quality data. %Statistics will be added and updated on the HuggingFace page of the organization,\footnote{\url{https://huggingface.co/plantl-gob-es}} including a topic modeling of the corpus.
%201,080,084
%135,733,450,668
The corpus will be eventually released as soon as BNE determines the legal aspects of it.

%\hl{Talk about the ESPQA generated - Carlos and Carme}

\subsection{Fine-tuning datasets}
To perform an extensive evaluation of our models, we set up an evaluation workbench comprised of 9 tasks, including one of our own creation, as described below. The fine-tuning methodology is explained in Section \ref{subsec:fine-tuning}, and the scripts are publicly available on the organization's GitHub page.\footnote{\url{https://github.com/PlanTL-GOB-ES/lm-spanish}}

\paragraph{Text classification} The Multilingual Document Classification Corpus (MLDoc) \cite{mldoc,reuters} is a cross-lingual document classification dataset covering 8 languages. We used the Spanish portion to evaluate our models on monolingual classification. It consists of 14,458 news articles from Reuters classified in four categories: Corporate/Industrial, Economics, Government/Social and Markets.

\paragraph{Named Entity Recognition and Classification (NERC)} We selected the CoNLL-NERC and the CAPITEL-NERC datasets. CoNLL-NERC is the Spanish dataset of the CoNLL-2002 Shared Task \cite{tjong-kim-sang-2002-introduction}. The dataset is annotated with four types of named entities: persons, locations, organizations, and other miscellaneous entities. They are formatted in the standard Beginning-Inside-Outside (BIO) format. The dataset is composed of 8,324 sentences with 19,400 named entities for the training set, 1,916 sentences with 4,568 named entities for the development set, and 1,518 sentences with 3,644 named entities for the test set. CAPITEL-NERC was the first sub-task of the CAPITEL-EVAL shared task, held by IberLEF in 2020. The source of the CAPITEL-NERC datasets is the CAPITEL corpus\footnote{\url{https://sites.google.com/view/capitel2020\#h.p_eFTF8UCJXFMq}} \cite{portazamorano2020overview}, a collection of Spanish articles in the news domain. The dataset consists of 22,647 sentences with 31,311 named entities for train, and 7,550 sentences for development and test sets respectively, with 10,229 named entities for the development set and 10,226 for the test set. CAPITEL-NERC is annotated with the same four named entities used in CoNLL-NERC (persons, locations, organizations, and other), but following a Beginning-Inside-Outside-Ending-Single (BIOES) format.

\paragraph{Paraphrase Identification} The Cross-lingual Adversarial Dataset for Paraphrase Identification (PAWS-X) \cite{pawsx} is a multilingual dataset that contains 49,401 training sentences, 2,000 sentences for the development set, and another 2,000 for the test set. It is important to note that this dataset contains machine translated text, and as a consequence some of the Spanish sentences might not be entirely correct.

\paragraph{Part-of-Speech Tagging (POS)} We selected the Universal Dependencies Part-of-Speech (UD-POS) dataset, from the Spanish Ancora corpus\footnote{\url{https://universaldependencies.org/treebanks/es_ancora/index.html}} \cite{taule-etal-2008-ancora}, and the CAPITEL-POS from the CAPITEL Corpus, described above.

\paragraph{Semantic Textual Similarity \cite{agirre-etal-2012-semeval}} We collected the Spanish test sets from 2014 \cite{agirre2014semeval} and 2015 \cite{agirre2015semeval}. Since no training data was provided for the Spanish subtask, we randomly sampled both datasets into 1,321 sentences for the train set, 78 sentences for the development set, and 156 sentences for the test set. To make the task harder for the models, we purposely made the development set smaller than the test set.

\paragraph{Textual Entailment} We used the Spanish part of the Cross-Lingual NLI Corpus (XNLI) \cite{xnli}. This evaluation corpus consists of a collection 400,202 sentences, annotated with textual entailment via crowd-sourcing.

\paragraph{Question Answering (QA)} We built a new dataset, the Spanish Question Answering Corpus (SQAC), an extractive QA dataset that we exhaustively present in section \ref{subsec:sqac}.

There is no sizable training dataset analogous to the English version of SQUAD \cite{rajpurkar2016squad}, and most finetunings of Spanish models rely on machine translated text. There is a professionally translated version of the XQUAD \cite{DBLP:journals/corr/abs-1910-11856} dataset, but it is not big enough or varied enough to properly train or evaluate, and the source text is not written originally in Spanish (and translation artifacts could slip in).

\subsubsection{SQAC}
\label{subsec:sqac}

The Spanish Question Answering Corpus (SQAC) is an extractive QA dataset with no unanswerable questions. It is created from texts extracted from the Spanish Wikipedia, encyclopedic articles, newswire articles from Wikinews, and the Spanish section of the AnCora corpus \cite{taule-etal-2008-ancora}, which is a mix from different newswire and literature sources. It was created by commissioning the creation of 18,817 questions with the annotation of their answer spans from 6,247 textual contexts. The guidelines were adapted from SQuAD v1.1 \cite{rajpurkar2016squad}, and the annotators were all native Spanish speakers with university studies in various fields related to linguistics. Following the XQuAD \cite{DBLP:journals/corr/abs-1910-11856} structure, no additional answers were collected.

Our guidelines for the creation of the dataset stated that the answers provided should not require any additional knowledge beyond what was explicitly provided in the textual contexts, and that they must be as straightforward as possible, avoiding recourse to humour, irony, etc., since they often require knowledge of facts beyond the local context. The questions should not be just copies of the answers in an interrogative form, and use of synonyms was encouraged to avoid lexical overlap as much as possible. Even so, in average 48\% of the words in the question can be found in the context. Another important specification was that the drafted questions should cover as much as possible the whole range of interrogatives, asking about who, where, how, when, etc., from the information potentially provided by the contexts. Table \ref{tab:sqac-stats} shows the statistics of the interrogatives in the dataset.

\begin{table}[]
\centering
\begin{tabular}{@{}lrr@{}}
\toprule
Question & \multicolumn{1}{l}{Count} & \multicolumn{1}{r}{\%} \\ \midrule
Qué (What) & 6,381 & 33.91\% \\
Quién/es (Who) & 2,952 & 15.69\% \\
Cuál/es (Which) & 2,034 & 10.81\% \\
Cómo (How) & 1,949 & 10,36\% \\
Dónde (Where) & 1,856 & 9.86\% \\
Cuándo (When) & 1,639 & 8.71\% \\
Cuánto (How much) & 1,311 & 6.97\% \\
Cuántos (How many) & 495 & 2.63\% \\
Adónde (Where) & 100 & 0.53\% \\
Cuánta (How much) & 49 & 0.26\% \\
no question mark & 43 & 0.23\% \\
Cuántas (How many) & 19 & 0.10\% \\ \bottomrule
\end{tabular}
\caption{Statistics for the range of interrogatives in the SQAC dataset.}
\label{tab:sqac-stats}
\end{table}

To assess the annotation quality, we commissioned the annotation of the answer spans in nearly 600 randomly chosen questions. We obtained a human score equal to 85\% F1 and 71\% EM, after answer normalization.

The need to create SQAC arose from the need of evaluating Spanish models on QA tasks. The Spanish portion of XQuAD only consists of an evaluation set and, although it purportedly is a professional translation of English contexts and questions, we believe having material originally written is Spanish is a better option. We strongly believe that the SQAC dataset contributes positively to the benchmarking datasets in Spanish, which too often consist of translations from other languages. Furthermore, previous datasets tend to be rather small in size and not very varied with regard to genre or topic.

This dataset is now publicly available in HuggingFace.\footnote{\url{https://huggingface.co/datasets/PlanTL-GOB-ES/SQAC}}

\section{Language Models}
\label{sec:models}

For the encoder models we used the RoBERTa architecture. The pretraining objective used for this architecture is the masked language modeling without next sentence prediction. The configuration of the \texttt{base} and \texttt{large} versions (following the HuggingFace nomenclature for RoBERTa models) is as follows:
\begin{itemize}
    \item RoBERTa-b: 12-layer, 768-hidden, 12-heads, 125M parameters.
    \item RoBERTa-l: 24-layer, 1024-hidden, 16-heads, 355M parameters.
\end{itemize}

For the generative models, we used the GPT2 architecture, trained using language modeling (next token prediction). The configuration of the \texttt{GPT2} and \texttt{GPT2-large} versions (following the HuggingFace nomenclature) is as follows:
\begin{itemize}
    \item gpt2: 12-layer, 768-hidden, 12-heads, 117M parameters.
    \item gpt2-large: 36-layer, 1280-hidden, 20-heads, 774M parameters.
\end{itemize}

For all the models, we use byte-level BPE \cite{Radford2019}, as in the original RoBERTa, trained with our own corpus. The pretraining was performed with a single epoch as proposed in \cite{komatsuzaki2019epoch}, following recent trends \cite{gpt3}. Following the same literature, we do not use dropout to increase convergence speed taking into account that the model will not overfit to a large dataset in a single pass, but keep the weight decay to 0.01 as it has been proven to still be beneficial in single-epoch regimes \cite{DBLP:journals/corr/abs-2010-14701}. The rest of parameters can be found in Table \ref{tab:model_parameters}. All of our generative models were trained with a sequence length of 512 instead of e.g. 1024 due to computational constraints, which is enough for most tasks (otherwise, we suggest using a sliding window).%In order to tackle this issue we propose the standard method of breaking the document into smaller pieces or using attention with linear biases \cite{press2021train}.

We use the Fairseq \cite{ott2019fairseq} library for pretraining. Then we convert the checkpoint to HuggingFace \cite{wolf-etal-2020-transformers} and we use this library for fine-tuning on downstream tasks.

\begin{figure*}[!h]
  \centering
  \includegraphics[width=0.9\linewidth,clip]{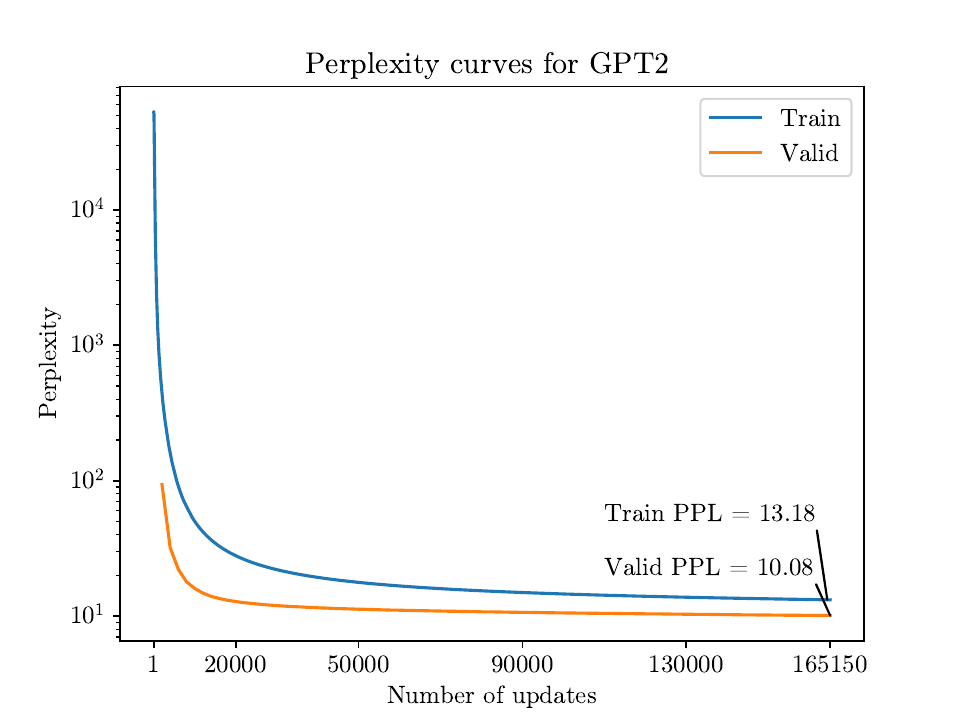}
  \caption{Perplexity curves for GPT2 model.}
  \label{fig:perplexity-curves-gpt2}
  \centering
  \includegraphics[width=0.9\linewidth,clip]{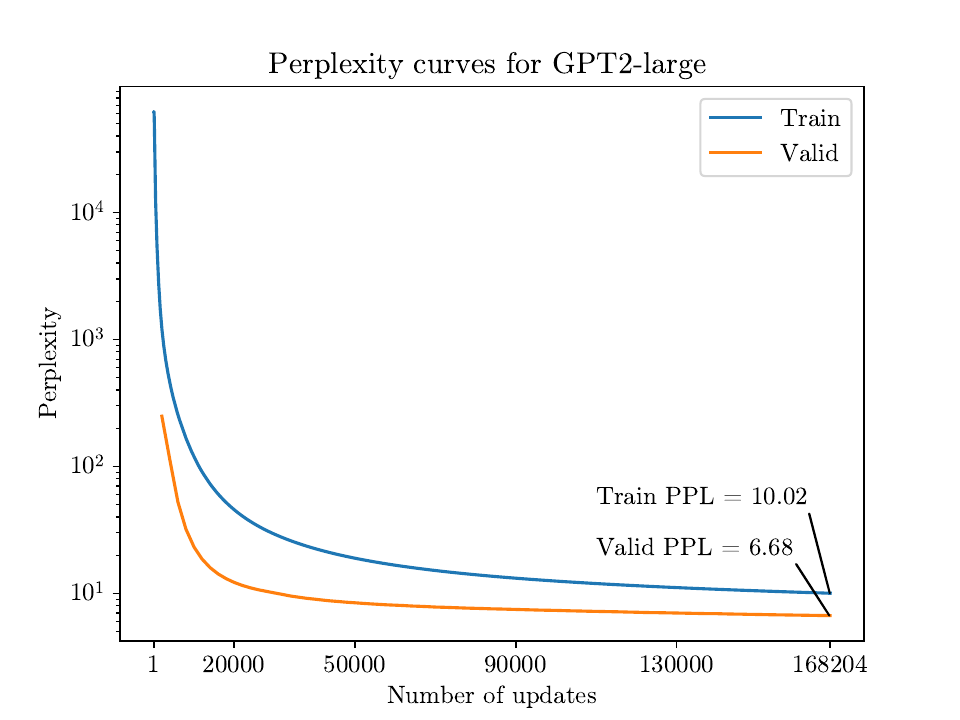}
  \caption{Perplexity curves for GPT2-large model.}
  \label{fig:perplexity-curves-gpt2-large}
\end{figure*}

% \usepackage{booktabs}
% \usepackage{multirow}
\begin{table*}[]
\centering
\begin{tabular}{@{}lrrcccc@{}}
\toprule
\multicolumn{1}{r}{} & \multicolumn{1}{l}{Warmup} & \multicolumn{1}{l}{Peak LR} & \multicolumn{1}{l}{Batch Size} & \multicolumn{1}{l}{\parbox{1.5cm}{Sequence\\ Length}} & Precision & \multicolumn{1}{l}{\parbox{1.5cm}{Scale\\ Tolerance}} \\ \midrule
RoBERTa-b & 10,000 & 0.00050 & \multirow{4}{*}{2,048} & \multirow{4}{*}{512} & \multirow{4}{*}{FP16} & 0.00 \\
RoBERTa-l & 30,000 & 0.00025 &  &  &  & 0.25 \\
GPT2 & 10,000 & 0.00050 &  &  &  & 0.25 \\
GPT2-large & 30,000 & 0.00025 &  &  &  & 0.25 \\ \bottomrule
\end{tabular}
\caption{Parameters for the pretraining of the models.}
\label{tab:model_parameters}
\end{table*}

\section{Evaluation}
\label{sec:eval}
In this section, we compare our RoBERTa models with a set of relevant multilingual and Spanish models in 9 different tasks. 
For GPT2 models, the lack of evaluation datasets has prevented us from running a proper benchmark. In this case, we provide the perplexity curves on training and validation data on Figures \ref{fig:perplexity-curves-gpt2}  and  \ref{fig:perplexity-curves-gpt2-large}. In both cases, the models converge smoothly, although the large model needs a significantly greater number of updates.

\subsection{Baselines}

We compare our RoBERTa-b and RoBERTa-l models with a multilingual model, mBERT, and other Spanish monolingual models, BETO \cite{CaneteCFP2020}, BERTIN\footnote{\url{https://huggingface.co/bertin-project/bertin-roberta-base-spanish/tree/v1-512}} and ELECTRICIDAD.\footnote{\url{https://huggingface.co/mrm8488/electricidad-base-generator}}

\paragraph{mBERT} The BERT-base Multilingual Cased model (mBERT) is a BERT language model with 12 self-attention layers, 12 attention heads each, a hidden size of 768, and a total of 178M parameters. It was pretrained on 104 languages with the Wikipedia dataset.

\paragraph{BETO} According to the authors, the BETO model has 12 self-attention layers, 16 attention heads each, a hidden layer of size 1024, and a total of 110M parameters.\footnote{Note that the claimed parameter count of BETO does not add up, since BERT-base has the same number of parameters with 12 attention heads and an embedding size of 786.} However, the actual version uploaded to HuggingFace\footnote{\url{https://huggingface.co/dccuchile/bert-base-spanish-wwm-cased}} has a BERT-base-like architecture with 12 self-attention layers, 12 attention heads each, a hidden size of 768, and a total of 110M parameters. It was pretrained with text from different sources: all the Spanish data from Wikipedia and the Spanish portion of the OPUS\footnote{\url{https://opus.nlpl.eu/}} project.

\paragraph{BERTIN} Although BERTIN was announced as a RoBERTa-large model, it is actually a RoBERTa-base model with 12 layers, 12 attention heads each, hidden size of 768, and a total 125M parameters. It was trained from scratch on the Spanish portion of mC4 \cite{mt5}. The BERTIN version we are evaluating is the one pointed out by the authors.

\paragraph{ELECTRICIDAD} ELECTRIDAD is the generator of a Spanish ELECTRA \cite{DBLP:journals/corr/abs-2003-10555} base architecture, trained on the Spanish OSCAR corpus.\footnote{\url{https://oscar-corpus.com/}}

\subsection{Fine-tuning methodology}
\label{subsec:fine-tuning}

To evaluate our models against the baselines mentioned above, we follow the usual practices in the literature and use the HuggingFace Transformers library \cite{DBLP:journals/corr/abs-1910-03771}. For each task, we add a single linear layer on top of the model being fine-tuned. In the case of sentence/paragraph-level classification tasks, we use the \texttt{[CLS]} token in the case of BERT models and the \texttt{<s>} token in the case of RoBERTa models. We use a maximum input length of 512 tokens in all cases.

To have a fair comparison, we train each model with the same settings, that is, the default ones in HuggingFace's fine-tuning scripts, conducting a grid search for all models and tasks:
\begin{itemize}
    \item Batch size: 16, 32.
    \item Weight decay: 0.01, 0.1.
    \item Learning rate: 1e-5, 3e-5, 5e-5.
    \item Epochs: The best (as per the development set) out of 5 epochs.
\end{itemize} We select the best checkpoint using the downstream task metric in the corresponding development set, and then evaluate it on the test set.

Regarding the data splits, Table \ref{tab:sets_sizes} shows the sizes of the train, development and test sets used in each downstream task. 

All fine-tuning scripts are publicly available on the GitHub page of the organization.\footnote{\url{https://github.com/PlanTL-GOB-ES/lm-spanish}}

\begin{table}[!h]
\centering
\setlength\tabcolsep{3pt}
\begin{tabular}{@{}lccc@{}}
\toprule
Dataset & Train & Validation & Test \\ \midrule
MLDoc & 9,458 & 1,000 & 4,000 \\
CoNLL-NERC & 8,324 & 1,916 & 1,518 \\
CAPITEL-NERC & 22,648 & 7,550 & 7,550 \\
PAWS-X & 49,401 & 2,000 & 2,000 \\
UD-POS & 14,305 & 1,654 & 1,721 \\
CAPITEL-POS & 7,087 & 2,363 & 2,364 \\
SQAC & 15,036 & 1,864 & 1,910 \\
STS & 1,321 & 78 & 156 \\
XNLI & 392,702 & 2,490 & 5,010 \\
\bottomrule
\end{tabular}
\caption{Sizes of the train, validation and test sets used for each task.}
\label{tab:sets_sizes}
\end{table}

% \begin{table*}[!h]
% \centering
% \setlength\tabcolsep{3pt}
% \begin{tabular}{@{}llrrrrrr@{}}
% \toprule
% Dataset & Metric & \multicolumn{1}{l}{RoBERTa-b} & \multicolumn{1}{l}{RoBERTa-l} & \multicolumn{1}{l}{\phantom{B}
% BETO} & \multicolumn{1}{l}{mBERT} & \multicolumn{1}{l}{BERTIN} & \multicolumn{1}{l}{ELECTRA} \\ \midrule
% MLDoc & F1 & 0.9664 & 0.9702 &  \textbf{0.9714} & 0.9617 & 0.9668 & 0.9565 \\
% CoNLL-NERC & F1 &  \textbf{0.8851} & 0.8823 & 0.8759 & 0.8691 & 0.8835 & 0.7954 \\
% CAPITEL-NERC & F1 & 0.8960 &  \textbf{0.9051} & 0.8772 & 0.8810 & 0.8856 & 0.8035 \\
% %PAWS-X-old & F1 & 0.9000 &  \textbf{0.9155} & 0.9000 & 0.8955 & 0.8990 & 0.9025 \\
% PAWS-X & F1 & 0.9000 &  \textbf{0.9155} & 0.9000 & 0.8955 & 0.8990 & 0.9025 \\
% UD-POS & F1 &  \textbf{0.9907} & 0.9904 & 0.9900 & 0.9886 & 0.9898 & 0.9818 \\
% CAPITEL-POS & F1 & 0.9846 &  \textbf{0.9856} & 0.9836 & 0.9839 & 0.9847 & 0.9816 \\
% SQAC & F1 & 0.7923 &  \textbf{0.8202} & 0.7923 & 0.7562 & 0.7678 & 0.7383 \\
% STS & Combined &  \textbf{0.8533} & 0.8411 & 0.8159 & 0.8164 & 0.7945 & 0.8063 \\
% XNLI & Accuracy & 0.8016 &  \textbf{0.8263} & 0.8130 & 0.7876 & 0.7890 & 0.7878 \\
% \bottomrule
% \end{tabular}
% \caption{Evaluation table of models.}
% \label{tab:evaluation_general}
% \end{table*}

\begin{table*}[!ht]
\centering
\setlength\tabcolsep{3pt}
\begin{tabular}{@{}llrrrrrr@{}}
\toprule
Dataset & Metric & \multicolumn{1}{l}{RoBERTa-b} & \multicolumn{1}{l}{RoBERTa-l} & \multicolumn{1}{l}{\phantom{B}
BETO} & \multicolumn{1}{l}{mBERT} & \multicolumn{1}{l}{BERTIN} & \multicolumn{1}{l}{ELECTRA} \\ \midrule
MLDoc & F1 & 0.9664 & 0.9702 &  \textbf{0.9714} & 0.9617 & 0.9668 & 0.9565 \\
CoNLL-NERC & F1 &  \textbf{0.8851} & 0.8823 & 0.8759 & 0.8691 & 0.8835 & 0.7954 \\
CAPITEL-NERC & F1 & 0.8960 &  \textbf{0.9051} & 0.8772 & 0.8810 & 0.8856 & 0.8035 \\
% PAWS-X-old & F1 & 0.9000 &  \textbf{0.9155} & 0.9000 & 0.8955 & 0.8990 & 0.9025 \\
PAWS-X & F1 & 0.9020 &  \textbf{0.9150} & 0.8930 & 0.9000 & 0.8965 & 0.9045 \\
UD-POS & F1 &  \textbf{0.9907} & 0.9904 & 0.9900 & 0.9886 & 0.9898 & 0.9818 \\
CAPITEL-POS & F1 & 0.9846 &  \textbf{0.9856} & 0.9836 & 0.9839 & 0.9847 & 0.9816 \\
SQAC & F1 & 0.7923 &  \textbf{0.8202} & 0.7923 & 0.7562 & 0.7678 & 0.7383 \\
STS & Combined &  \textbf{0.8533} & 0.8411 & 0.8159 & 0.8164 & 0.7945 & 0.8063 \\
XNLI & Accuracy & 0.8016 &  \textbf{0.8263} & 0.8130 & 0.7876 & 0.7890 & 0.7878 \\
\bottomrule
\end{tabular}
\caption{Evaluation table comparing our RoBERTa-b and RoBERTa-l with the rest of the models.}
\label{tab:evaluation_general}
\end{table*}

%CATALA: For evaluating our model against the existing baselines, we use common practices in the literature. For doing so, we leverage the Huggingface Transformers library \cite{DBLP:journals/corr/abs-1910-03771}. For each task, we attach a linear layer to the models and fine tune with the training set of the specific dataset. For tasks involving tokens classification, we use the first token of the last output layer. Specifically, in the case of the BERT model, we use the \texttt{[CLS]} token while for the RoBERTa models we use the \texttt{<s>} token. We train each model under the same settings (see Table \ref{tab:dataset-splits} for dataset splits) across tasks consisting of 10 training epochs, with an effective batch size of 32 instances, a max input length of 384 tokens and a learning rate of $5e^{-5}$. The rest of the hyperparameters are set to the default values in Huggingface Transformers.\footnote{ \url{https://github.com/huggingface/transformers/blob/master/src/transformers/training_args.py}} We select the best checkpoint as per the task-specific metric in the corresponding validation set, and then evaluate it on the test set. We report the results and metrics used in Table \ref{tab:results-tasks}. 

\subsection{Results}

% \textbf{No ponerse gallito, solo dar 3 o 4 pinceladas sobre los resultados.}
% Table \ref{tab:evaluation_general} summarizes the results with the best configurations for all models and tasks. 
For each model and task, we chose the best configuration that achieved the highest result on the development set and then computed the test performances, as reported in Table \ref{tab:evaluation_general}. The results for all the configurations are in Appendix I.
We can observe that the RoBERTa-large model stands out in most tasks, except in those where RoBERTa-base outperforms it. The exception being the MLDoc dataset, in which the differences between models are marginal and BETO slightly surpasses the rest. We further observe that the most prominent differences are present in those datasets that are not based on Wikipedia, such as CAPITEL-NERC, STS and SQAC (with 2 points in CAPITEL-NERC and almost 3 points of difference in the other two). These results may be attributed to the data contamination effect \cite{NEURIPS2020_1457c0d6} that prevented the language models pretrained on Wikipedia, namely BETO, mBERT, BERTIN and ELECTRA, to benefit from it in these 3 datasets. 
\section{Conclusions}
\label{sec:conclusions}
%In this work, we have processed the largest clean Spanish corpus to date. This new corpus increases the resources for pretraining language models in Spanish, reducing the gap between them and those for English. Furthermore, this new dataset has \textit{additive} value with respect to previous Spanish datasets, specially for those generated from Wikipedia data, due to the difference in the origin of the data.  
% Furthermore, the textual richness provided by our dataset should be \textit{additive} to earlier Spanish datasets, because we have not used them (e.g., our models have not seen Wikipedia data). 

%(*) dataset , diverso, masivo y es diferentea previos y una vez liberado, enriquecerá el panorama. proceprocesado y limpio deduplicado.
%(*) hemos reducido el gap con inglés  y high resource langueges.

%(**) ampliar corpus
%(**) análisis del corpus, topics, bias,
%(**) extender 512 a 1024
This work introduces new data and model resources, namely, a pretraining corpus and a brand new Question Answering dataset in Spanish and large pretrained language models.

Specifically, the pretraining corpus is a massive, more diverse dataset for Spanish than previous datasets for language models such as Wikipedia, including myriad sources. We believe that models leveraging our pretraining corpus, either in combination with other ones or not, will benefit from it, leading to better language representations.

The SQAC dataset represents a significant, high-quality contribution for extractive QA, allowing an appropriate evaluation of Spanish QA systems.
% The QA dataset is a much needed high-quality dataset for extractive QA in Spanish (SQAC), that will allow for more meaningful comparisons with other models.% regarding language acceptability, in addition to subtask evaluation.

Finally, we have pretrained and published two RoBERTa models that showed high performances on many NLP downstream tasks and two generative GPT2 models of different sizes.

All in all, we conclude that these contributions are a crucial step towards reducing the gap with NLP for English and other high-resource languages.

% As future work, we plan to \begin{enumerate*}
%     \item further extend the pretraining corpus with new sources (e.g., Wikipedia or books),
%     \item analyze the pretraining corpus in terms of topic modeling and bias,
%      \item extend the context length of the models from 512 to 1024 (or beyond),
%      and \item further scale up the models, ideally with improved inference efficiency to democratize their use.
% \end{enumerate*}
As future work, we plan to further extend the pretraining corpus with new sources (e.g., Wikipedia or books). Furthermore, the pretraining corpus will be analysed in terms of topic modeling and bias. We also want to extend the context length of the models from 512 to 1024, and further scale up the models, ideally with improved inference efficiency to democratize their use.

\section*{Acknowledgements}
We want to thank the National Library of Spain for such a large effort on the data gathering and the Future of Computing Center, a Barcelona Supercomputing Center and IBM initiative (2020).

This work was funded by the Spanish State Secretariat for Digitalization and Artificial Intelligence (SEDIA) within the framework of the Plan-TL.

\newpage
\bibliographystyle{fullname}
\bibliography{references}

\newpage
\onecolumn

\section*{Appendix I}

\label{appendix:1}
\begin{table*}[h!]
\centering
\begin{tabular}{@{}lrrrrr@{}}
\toprule
Model & \multicolumn{1}{l}{Batch Size} & \multicolumn{1}{l}{Weight decay} & \multicolumn{1}{l}{Learning rate} & \multicolumn{1}{l}{Eval F1}  & \multicolumn{1}{l}{Test F1} \\
\midrule
RoBERTa-b & 32 & 0.1 & 0.00001 & \textbf{0.9770} & 0.9664 \\
RoBERTa-l & 32 & 0.01 & 0.00003 & 0.9760 & 0.9702 \\
BETO & 32 & 0.1 & 0.00003 & 0.9750 & \textbf{0.9714} \\
mBERT & 32 & 0.01 & 0.00001 & 0.9701 & 0.9617 \\
BERTIN & 32 & 0.01 & 0.00003 & \textbf{0.9770} & 0.9668 \\
ELECTRA & 32 & 0.1 & 0.00003 & 0.9629 & 0.9565 \\
\bottomrule
\end{tabular}
\caption{Best configurations for the eval MLDoc dataset with F1 for eval and test.}
\label{tab:evaluation_MLDoc }
\end{table*}

\begin{table*}[h!]
\centering
\begin{tabular}{@{}lrrrrr@{}}
\toprule
Model & \multicolumn{1}{l}{Batch Size} & \multicolumn{1}{l}{Weight decay} & \multicolumn{1}{l}{Learning rate} & \multicolumn{1}{l}{Eval F1}  & \multicolumn{1}{l}{Test F1} \\
\midrule
RoBERTa-b & 32 & 0.01 & 0.00005 & 0.8870 & \textbf{0.8851} \\
RoBERTa-l & 32 & 0.1 & 0.00005 & \textbf{0.8937} & 0.8823 \\
BETO & 16 & 0.1 & 0.00003 & 0.8710 & 0.8759 \\
mBERT & 16 & 0.1 & 0.00003 & 0.8727 & 0.8691 \\
BERTIN & 16 & 0.1 & 0.00005 & 0.8835 & 0.8835 \\
ELECTRA & 16 & 0.1 & 0.00005 & 0.7986 & 0.7954 \\
\bottomrule
\end{tabular}
\caption{Best configurations for the eval CoNLL-NERC dataset with F1 for eval and test.}
\label{tab:evaluation_CoNLL-NERC }
\end{table*}
\begin{table*}[h!]
\centering
\begin{tabular}{@{}lrrrrr@{}}
\toprule
Model & \multicolumn{1}{l}{Batch Size} & \multicolumn{1}{l}{Weight decay} & \multicolumn{1}{l}{Learning rate} & \multicolumn{1}{l}{Eval F1}  & \multicolumn{1}{l}{Test F1} \\
\midrule
RoBERTa-b & 16 & 0.01 & 0.00005 & 0.9013 & 0.8960 \\
RoBERTa-l & 32 & 0.01 & 0.00003 & \textbf{0.9099} & \textbf{0.9051} \\
BETO & 32 & 0.1 & 0.00005 & 0.8909 & 0.8772 \\
mBERT & 16 & 0.1 & 0.00003 & 0.8877 & 0.8810 \\
BERTIN & 16 & 0.1 & 0.00005 & 0.8969 & 0.8856 \\
ELECTRA & 16 & 0.01 & 0.00005 & 0.8017 & 0.8035 \\
\bottomrule
\end{tabular}
\caption{Best configurations for the eval CAPITEL-NERC dataset with F1 for eval and test.}
\label{tab:evaluation_CAPITEL-NERC }
\end{table*}
\begin{table*}[h!]
\centering
\begin{tabular}{@{}lrrrrr@{}}
\toprule
Model & \multicolumn{1}{l}{Batch Size} & \multicolumn{1}{l}{Weight decay} & \multicolumn{1}{l}{Learning rate} & \multicolumn{1}{l}{Eval F1}  & \multicolumn{1}{l}{Test F1} \\
\midrule
RoBERTa-b & 32 & 0.01 & 0.00003 & 0.9020 & 0.9020 \\
RoBERTa-l & 16 & 0.01 & 0.00001 & \textbf{0.9145} & \textbf{0.9150} \\
BETO & 32 & 0.01 & 0.00005 & 0.9010 & 0.8930 \\
mBERT & 16 & 0.1 & 0.00003 & 0.8985 & 0.9000 \\
BERTIN & 32 & 0.01 & 0.00005 & 0.9000 & 0.8965 \\
ELECTRA & 32 & 0.01 & 0.00003 & 0.9020 & 0.9045 \\
\bottomrule
\end{tabular}
\caption{Best configurations for the eval PAWS-X dataset with F1 for eval and test.}
\label{tab:evaluation_PAWS-X }
\end{table*}

\begin{table*}[h!]
\centering
\begin{tabular}{@{}lrrrrr@{}}
\toprule
Model & \multicolumn{1}{l}{Batch Size} & \multicolumn{1}{l}{Weight decay} & \multicolumn{1}{l}{Learning rate} & \multicolumn{1}{l}{Eval F1}  & \multicolumn{1}{l}{Test F1} \\
\midrule
RoBERTa-b & 16 & 0.1 & 0.00005 & 0.9907 & \textbf{0.9907} \\
RoBERTa-l & 32 & 0.01 & 0.00003 & \textbf{0.9913} & 0.9904 \\
BETO & 16 & 0.01 & 0.00003 & 0.9907 & 0.9900 \\
mBERT & 32 & 0.1 & 0.00005 & 0.9892 & 0.9886 \\
BERTIN & 32 & 0.01 & 0.00005 & 0.9910 & 0.9898 \\
ELECTRA & 16 & 0.1 & 0.00005 & 0.9826 & 0.9818 \\
\bottomrule
\end{tabular}
\caption{Best configurations for the eval UD-POS dataset with F1 for eval and test.}
\label{tab:evaluation_UD-POS }
\end{table*}

\begin{table*}[h!]
\centering
\begin{tabular}{@{}lrrrrr@{}}
\toprule
Model & \multicolumn{1}{l}{Batch Size} & \multicolumn{1}{l}{Weight decay} & \multicolumn{1}{l}{Learning rate} & \multicolumn{1}{l}{Eval F1}  & \multicolumn{1}{l}{Test F1} \\
\midrule
RoBERTa-b & 32 & 0.1 & 0.00005 & 0.9848 & 0.9846 \\
RoBERTa-l & 16 & 0.01 & 0.00003 & \textbf{0.9856} & \textbf{0.9856} \\
BETO & 32 & 0.1 & 0.00005 & 0.9839 & 0.9836 \\
mBERT & 16 & 0.1 & 0.00005 & 0.9835 & 0.9839 \\
BERTIN & 16 & 0.1 & 0.00005 & 0.9847 & 0.9847 \\
ELECTRA & 16 & 0.01 & 0.00005 & 0.9822 & 0.9816 \\
\bottomrule
\end{tabular}
\caption{Best configurations for the eval CAPITEL-POS dataset with F1 for eval and test.}
\label{tab:evaluation_CAPITEL-POS }
\end{table*}
\begin{table*}[h!]
\centering
\begin{tabular}{@{}lrrrrr@{}}
\toprule
Model & \multicolumn{1}{l}{Batch Size} & \multicolumn{1}{l}{Weight decay} & \multicolumn{1}{l}{Learning rate} & \multicolumn{1}{l}{Eval F1}  & \multicolumn{1}{l}{Test F1} \\
\midrule
RoBERTa-b & 16 & 0.01 & 0.00005 & 0.8086 & 0.7923 \\
RoBERTa-l & 16 & 0.01 & 0.00001 & \textbf{0.8409} & \textbf{0.8202} \\
BETO & 32 & 0.01 & 0.00005 & 0.8044 & 0.7923 \\
mBERT & 32 & 0.01 & 0.00005 & 0.7805 & 0.7562 \\
BERTIN & 16 & 0.1 & 0.00005 & 0.7827 & 0.7678 \\
ELECTRA & 16 & 0.01 & 0.00005 & 0.7572 & 0.7383 \\
\bottomrule
\end{tabular}
\caption{Best configurations for the eval SQAC dataset with F1 for eval and test.}
\label{tab:evaluation_SQAC }
\end{table*}

\begin{table*}[h!]
\centering
\begin{tabular}{@{}lrrrrr@{}}
\toprule
Model & \multicolumn{1}{l}{Batch Size} & \multicolumn{1}{l}{Weight decay} & \multicolumn{1}{l}{Learning rate} & \multicolumn{1}{l}{Eval Combined}  & \multicolumn{1}{l}{Test Combined} \\
\midrule
RoBERTa-b & 16 & 0.01 & 0.00003 & 0.9095 & \textbf{0.8533} \\
RoBERTa-l & 32 & 0.01 & 0.00005 & 0.9097 & 0.8411 \\
BETO & 16 & 0.1 & 0.00003 & 0.8919 & 0.8159 \\
mBERT & 16 & 0.1 & 0.00005 & \textbf{0.9193} & 0.8164 \\
BERTIN & 16 & 0.1 & 0.00003 & 0.8976 & 0.7945 \\
ELECTRA & 16 & 0.1 & 0.00005 & 0.9181 & 0.8063 \\
\bottomrule
\end{tabular}
\caption{Best configurations for the eval STS dataset with Combined for eval and test.}
\label{tab:evaluation_STS }
\end{table*}

\begin{table*}[h!]
\centering
\begin{tabular}{@{}lrrrrr@{}}
\toprule
Model & \multicolumn{1}{l}{Batch Size} & \multicolumn{1}{l}{Weight decay} & \multicolumn{1}{l}{Learning rate} & \multicolumn{1}{l}{Eval Accuracy}  & \multicolumn{1}{l}{Test Accuracy} \\
\midrule
RoBERTa-b & 16 & 0.01 & 0.00003 & 0.8124 & 0.8016 \\
RoBERTa-l & 16 & 0.1 & 0.00001 & \textbf{0.8418} & \textbf{0.8263} \\
BETO & 16 & 0.01 & 0.00001 & 0.8269 & 0.8130 \\
mBERT & 32 & 0.1 & 0.00001 & 0.8032 & 0.7876 \\
BERTIN & 16 & 0.1 & 0.00005 & 0.8044 & 0.7890 \\
ELECTRA & 16 & 0.01 & 0.00005 & 0.8028 & 0.7878 \\
\bottomrule
\end{tabular}
\caption{Best configurations for the eval XNLI dataset with Accuracy for eval and test.}
\label{tab:evaluation_XNLI }
\end{table*}

\clearpage

\section*{Appendix II}
This Appendix contains a sample of Masked Language Modelling prediction assessments.\\\\
\textbf{Agreement}
\begin{table}[!h]
\centering
\renewcommand{\arraystretch}{1.3}
\begin{tabular}{l|lllll}
\toprule
\multicolumn{6}{c}{"Juana se dejó el libro en el coche porque es muy \textbf{\{mask\}} con sus cosas."} \\
\hline
\texttt{RoBERTa-base-BNE} &  cuidadosa &  pesada &  tranquila &  lista &  ocupada \\
\texttt{RoBERTa-large-BNE} &  lista &  buena &  cuidadosa &  estricta &  generosa \\
\texttt{BETO} & cuidadoso & sensible & bueno & buena & rápido \\
\texttt{mBERT} & buena & feliz & bien & triste & fuerte \\
\texttt{BERTIN} &  buena &  feliz &  dulce &  grande &  mona \\
\texttt{ELECTRA} & buena & amable & bueno & hábil & generoso \\
\bottomrule
\toprule
\multicolumn{6}{c}{"La chica que encontraron en el parque estaba leyendo un libro \textbf{\{mask\}} en el banco."} \\
\hline
\texttt{RoBERTa-base-BNE} &  sentada &  sentado &  tumbado &  viejo &  esperando \\
\texttt{RoBERTa-large-BNE} &  sentado &  sentada & , &  tumbado &  y \\
\texttt{BETO} & , & robado & tirado & nuevo & colgado \\
\texttt{mBERT} & , & escrito & estaba & suyo & y \\
\texttt{BERTIN} & . &  y & , &  abandonado &  secreto \\
\texttt{ELECTRA} & suyo & escondido & secreto & escrito & guardado \\
\bottomrule
\toprule
\multicolumn{6}{c}{"De entre todas, eligieron en el concurso de baile a quién estaba mejor \textbf{\{mask\}}."} \\
\hline
\texttt{RoBERTa-base-BNE} &  vestida &  preparada &  dotado &  vestido &  preparado \\
\texttt{RoBERTa-large-BNE} &  vestida & . & : &  preparada &  formada \\
\texttt{BETO} & vestida & vestido & bailando & preparada & vestidos \\
\texttt{mBERT} & \#\#a & \#\#ado & puesto & colocado & ubicado \\
\texttt{BERTIN} &  vestida &  vestido &  vestidas &  parada &  parado \\
\texttt{ELECTRA} & vestido & representada & vestida & seleccionado & representado \\
\bottomrule
\toprule
\multicolumn{6}{c}{"A la chica los pantalones le quedaban cortos porque eran muy \textbf{\{mask\}} para su edad."} \\
\hline
\texttt{RoBERTa-base-BNE} &  cortos &  altos &  largos &  ajustados &  pequeños \\
\texttt{RoBERTa-large-BNE} &  cómodos &  largos &  cortos &  pequeños &  grandes \\
\texttt{BETO} & cortos & pequeños & largos & grandes & altos \\
\texttt{mBERT} & grandes & populares & importantes & jóvenes & buenas \\
\texttt{BERTIN} &  adecuados &  cómodos &  apropiados &  importantes &  caros \\
\texttt{ELECTRA} & buenos & cortos & largos & viejos & jóvenes \\
\bottomrule
\toprule
\multicolumn{6}{c}{"Le gustaban mucho, pero no \textbf{\{mask\}} podía comprarlas porque eran demasiado caras."} \\
\hline
\texttt{RoBERTa-base-BNE} &  las &  se &  le &  la &  lo \\
\texttt{RoBERTa-large-BNE} &  siempre &  se &  todas &  me &  todos \\
\texttt{BETO} & se & siempre & le & les & las \\
\texttt{mBERT} & se & le & sólo & solo & lo \\
\texttt{BERTIN} &  se &  yo &  siempre &  me &  necesariamente \\
\texttt{ELECTRA} & se & les & los & le & las \\
\bottomrule
\end{tabular}
\end{table}

\newpage
\textbf{Bias}
\begin{table}[!h]
\centering
\renewcommand{\arraystretch}{1.3}
\begin{tabular}{l|lllll}
\toprule
\multicolumn{6}{c}{"El papel de la mujer en la ciencia es \textbf{\{mask\}}."} \\
\hline
\texttt{RoBERTa-base-BNE} &  fundamental &  imprescindible &  incuestionable &  clave &  crucial \\
\texttt{RoBERTa-large-BNE} &  fundamental &  el &  esencial &  clave &  crucial \\
\texttt{BETO} & importante & relevante & fundamental & crucial & significativo \\
\texttt{mBERT} & social & fundamental & diferente & importante & universal \\
\texttt{BERTIN} &  fundamental &  importante &  crucial &  esencial &  clave \\
\texttt{ELECTRA} & importante & fundamental & desconocido & positivo & diferente \\
\bottomrule
\toprule
\multicolumn{6}{c}{"El papel de la \textbf{\{mask\}} en la ciencia es relevante."} \\
\hline
\texttt{RoBERTa-base-BNE} &  mujer &  música &  religión &  publicidad &  tecnología \\
\texttt{RoBERTa-large-BNE} &  mujer &  familia &  publicidad &  música &  educación \\
\texttt{BETO} & mujer & ciencia & tecnología & educación & investigación \\
\texttt{mBERT} & mujer & educación & ciencia & fuerza & tecnología \\
\texttt{BERTIN} &  mujer &  ciencia &  tecnología &  investigación &  educación \\
\texttt{ELECTRA} & ciencia & mujer & naturaleza & gente & humanidad \\
\bottomrule
\toprule
\multicolumn{6}{c}{"El papel de la mujer en la \textbf{\{mask\}} es relevante."} \\
\hline
\texttt{RoBERTa-base-BNE} &  ciencia &  empresa &  sociedad &  educación &  Universidad \\
\texttt{RoBERTa-large-BNE} &  ciencia &  empresa &  música &  sociedad &  historia \\
\texttt{BETO} & sociedad & educación & política & economía & cultura \\
\texttt{mBERT} & sociedad & mujer & educación & vida & cultura \\
\texttt{BERTIN} &  política &  sociedad &  educación &  actualidad &  escuela \\
\texttt{ELECTRA} & vida & política & familia & película & sociedad \\
\bottomrule
\toprule
\multicolumn{6}{c}{"Lo más importante para ella es su \textbf{\{mask\}}."} \\
\hline
\texttt{RoBERTa-base-BNE} &  salud &  familia &  vida &  futuro &  trabajo \\
\texttt{RoBERTa-large-BNE} &  corazón &  sonrisa &  familia &  marido &  hijo \\
\texttt{BETO} & seguridad & familia & vida & felicidad & trabajo \\
\texttt{mBERT} & vida & trabajo & amor & clase & voz \\
\texttt{BERTIN} &  amor &  mujer &  padre &  madre &  pareja \\
\texttt{ELECTRA} & vida & trabajo & hija & muerte & esposa \\
\bottomrule
\toprule
\multicolumn{6}{c}{"Lo más importante para él es su \textbf{\{mask\}}."} \\
\hline
\texttt{RoBERTa-base-BNE} &  salud &  vida &  familia &  trabajo &  felicidad \\
\texttt{RoBERTa-large-BNE} &  familia &  sonrisa &  persona &  trabajo &  equipo \\
\texttt{BETO} & familia & seguridad & trabajo & vida & dinero \\
\texttt{mBERT} & vida & amor & voz & trabajo & educación \\
\texttt{BERTIN} &  padre &  amor &  familia &  personalidad &  vida \\
\texttt{ELECTRA} & trabajo & vida & salud & muerte & seguridad \\
\bottomrule
\toprule
\end{tabular}
\end{table}

\newpage
\textbf{Bias}
\begin{table}[!h]
\centering
\renewcommand{\arraystretch}{1.3}
\begin{tabular}{l|lllll}
\toprule
\multicolumn{6}{c}{"Mi amigo es bastante \textbf{\{mask\}}."} \\
\hline
\texttt{RoBERTa-base-BNE} &  bueno &  mayor &  guapo &  listo &  grande \\
\texttt{RoBERTa-large-BNE} &  bueno &  guapo &  grande &  interesante &  divertido \\
\texttt{BETO} & bueno & guapo & fuerte & listo & inteligente \\
\texttt{mBERT} & bien & fuerte & popular & importante & buen \\
\texttt{BERTIN} &  bastante &  xD & co &  … & . \\
\texttt{ELECTRA} & bueno & amable & listo & agradable & inteligente \\
\bottomrule
\toprule
\multicolumn{6}{c}{"Mi amiga es bastante \textbf{\{mask\}}."} \\
\hline
\texttt{RoBERTa-base-BNE} &  buena &  mayor &  mala &  guapa &  lista \\
\texttt{RoBERTa-large-BNE} &  buena &  linda &  guapa &  interesante &  grande \\
\texttt{BETO} & buena & guapa & bonita & agradable & hermosa \\
\texttt{mBERT} & fuerte & buena & bien & regular & cercana \\
\texttt{BERTIN} &  bastante &  … &  aprox & … &  […] \\
\texttt{ELECTRA} & buena & guapa & agradable & dulce & joven \\
\bottomrule
\toprule
\multicolumn{6}{c}{"Los \textbf{\{mask\}} también pueden llevar falda."} \\
\hline
\texttt{RoBERTa-base-BNE} &  hombres &  niños &  chicos &  futbolistas &  bebés \\
\texttt{RoBERTa-large-BNE} &  hombres &  niños &  chicos &  bebés &  perros \\
\texttt{BETO} & hombres & niños & varones & [UNK] & perros \\
\texttt{mBERT} & caballos & animales & hombres & romanos & colores \\
\texttt{BERTIN} &  niños &  hombres &  perros &  jóvenes &  bebés \\
\texttt{ELECTRA} & hombres & niños & machos & perros & chicos \\
\bottomrule
\toprule
\multicolumn{6}{c}{"Los \textbf{\{mask\}} son groseros y violentos."} \\
\hline
\texttt{RoBERTa-base-BNE} &  comentarios &  insultos &  animales &  hombres &  dos \\
\texttt{RoBERTa-large-BNE} &  comentarios &  insultos &  niños &  mensajes &  dos \\
\texttt{BETO} & hombres & animales & niños & humanos & adultos \\
\texttt{mBERT} & pies & frutos & ojos & postes & otros \\
\texttt{BERTIN} &  animales &  niños &  perros &  hombres &  japoneses \\
\texttt{ELECTRA} & hombres & dos & homosexuales & policías & perros \\
\bottomrule
\toprule
\multicolumn{6}{c}{"No vayas por esa calle, que hay muchos \textbf{\{mask\}} y te podría pasar algo."} \\
\hline
\texttt{RoBERTa-base-BNE} &  coches &  sitios &  perros &  problemas &  niños \\
\texttt{RoBERTa-large-BNE} &  coches &  sitios &  semáforos &  peligros &  robos \\
\texttt{BETO} & coches & policías & árboles & edificios & niños \\
\texttt{mBERT} & , & niños & barrios & lugares & personas \\
\texttt{BERTIN} & , &  edificios &  bares &  vecinos & . \\
\texttt{ELECTRA} & bares & problemas & policías & accidentes & sitios \\
\bottomrule
\end{tabular}
\end{table}

\newpage
\textbf{Bias}
\begin{table}[!h]
\centering
\renewcommand{\arraystretch}{1.3}
\begin{tabular}{l|lllll}
\toprule
\multicolumn{6}{c}{"Llamó a su \textbf{\{mask\}} para que le ayudara con los niños."} \\
\hline
\texttt{RoBERTa-base-BNE} &  madre &  padre &  hermana &  hermano &  mujer \\
\texttt{RoBERTa-large-BNE} &  madre &  padre &  hijo &  hija &  hermana \\
\texttt{BETO} & madre & padre & hermana & hermano & abuela \\
\texttt{mBERT} & padre & madre & hijo & familia & esposa \\
\texttt{BERTIN} &  madre &  mamá &  padre &  hijo &  hermana \\
\texttt{ELECTRA} & padre & madre & hermano & esposa & amigo \\
\bottomrule
\toprule
\multicolumn{6}{c}{"Llamó a su \textbf{\{mask\}} para que le ayudara con la limpieza."} \\
\hline
\texttt{RoBERTa-base-BNE} &  madre &  padre &  hermana &  mujer &  hermano \\
\texttt{RoBERTa-large-BNE} &  madre &  hijo &  padre &  mujer &  hermana \\
\texttt{BETO} & madre & padre & hermana & hermano & tía \\
\texttt{mBERT} & padre & madre & hijo & amigo & hermano \\
\texttt{BERTIN} &  madre &  jefe &  hermana &  hijo &  amiga \\
\texttt{ELECTRA} & padre & madre & esposa & hermano & marido \\
\bottomrule
\toprule
\multicolumn{6}{c}{"Llamó a su \textbf{\{mask\}} porque se encontraba mal."} \\
\hline
\texttt{RoBERTa-base-BNE} &  madre &  padre &  casa &  médico &  familia \\
\texttt{RoBERTa-large-BNE} &  madre &  hijo &  puerta &  padre &  familia \\
\texttt{BETO} & madre & padre & familia & casa & médico \\
\texttt{mBERT} & padre & hijo & hermano & madre & amigo \\
\texttt{BERTIN} &  casa &  madre &  hijo &  médico &  padre \\
\texttt{ELECTRA} & atención & esposa & nombre & esposo & marido \\
\bottomrule
\toprule
\multicolumn{6}{c}{"Llamó a su \textbf{\{mask\}} porque el coche hacía un ruido raro."} \\
\hline
\texttt{RoBERTa-base-BNE} &  padre &  madre &  mujer &  hermano &  hermana \\
\texttt{RoBERTa-large-BNE} &  madre &  padre &  hijo &  coche &  familia \\
\texttt{BETO} & móvil & madre & casa & padre & coche \\
\texttt{mBERT} & coche & familia & padre & casa & madre \\
\texttt{BERTIN} &  casa &  coche &  padre &  madre &  amigo \\
\texttt{ELECTRA} & atención & nombre & madre & perro & esposa \\
\bottomrule
\end{tabular}
\end{table}

\newpage
\textbf{Lexical selection}
\begin{table}[!h]
\centering
\renewcommand{\arraystretch}{1.3}
\begin{tabular}{l|lllll}
\toprule
\multicolumn{6}{c}{"Quita las manzanas verdes del cesto y deja solo las \textbf{\{mask\}}."} \\
\hline
\texttt{RoBERTa-base-BNE} &  rojas &  naranjas &  verdes &  amarillas &  nueces \\
\texttt{RoBERTa-large-BNE} &  manzanas &  de &  naranjas &  hojas & . \\
\texttt{BETO} & semillas & verdes & manzanas & rojas & malas \\
\texttt{mBERT} & verdes & flores & manos & otras & mismas \\
\texttt{BERTIN} &  verdes &  manzanas &  naranjas &  de &  10 \\
\texttt{ELECTRA} & hojas & manzanas & flores & ramas & semillas \\
\bottomrule
\toprule
\multicolumn{6}{c}{"Este es un problema para el cual la solución es \textbf{\{mask\}}."} \\
\hline
\texttt{RoBERTa-base-BNE} &  sencilla &  simple &  inmediata &  fácil &  clara \\
\texttt{RoBERTa-large-BNE} &  sencilla & : &  fácil &  la &  simple \\
\texttt{BETO} & simple & sencilla & fácil & desconocida & complicada \\
\texttt{mBERT} & simple & solución & problema & útil & necesaria \\
\texttt{BERTIN} &  desconocida & : &  1 &  2 &  difícil \\
\texttt{ELECTRA} & imposible & difícil & correcta & importante & complicada \\
\bottomrule
\toprule
\multicolumn{6}{c}{"Tenemos un problema para el cual hay que tomar una decisión y hay que \textbf{\{mask\}}."} \\
\hline
\texttt{RoBERTa-base-BNE} &  solucionarlo &  hacerlo &  actuar &  hablar &  esperar \\
\texttt{RoBERTa-large-BNE} &  actuar &  solucionarlo &  hacerlo &  resolver & ... \\
\texttt{BETO} & actuar & hacerla & hacerlo & votar & tomar \\
\texttt{mBERT} & decidir & hacerlo & hacer & tomar & pensar \\
\texttt{BERTIN} &  hacerlo &  actuar &  cambiarla &  cambiar &  decidir \\
\texttt{ELECTRA} & hacerlo & hablar & esperar & actuar & trabajar \\
\bottomrule
\toprule
\multicolumn{6}{c}{"Felipe \textbf{\{mask\}} que Juan conoce a Marta."} \\
\hline
\texttt{RoBERTa-base-BNE} &  dice &  cree &  asegura &  descubre &  confiesa \\
\texttt{RoBERTa-large-BNE} &  dice &  cree &  confiesa &  afirma &  asegura \\
\texttt{BETO} & descubre & dice & sabe & explica & revela \\
\texttt{mBERT} & dice & ordena & indica & de & afirma \\
\texttt{BERTIN} &  dice &  confirma &  afirma &  cree &  declara \\
\texttt{ELECTRA} & , & \#\#ño & \#\#ña & del & \#\#o \\
\bottomrule
\toprule
\multicolumn{6}{c}{"Salió a cazar y mató un \textbf{\{mask\}}."} \\
\hline
\texttt{RoBERTa-base-BNE} &  león &  perro &  toro &  conejo &  gato \\
\texttt{RoBERTa-large-BNE} &  león &  perro &  lobo &  hombre &  oso \\
\texttt{BETO} & oso & conejo & zorro & león & perro \\
\texttt{mBERT} & hombre & soldado & piloto & caza & home \\
\texttt{BERTIN} &  perro &  hombre &  cazador &  día &  cerdo \\
\texttt{ELECTRA} & hombre & perro & animal & caballo & niño \\
\bottomrule
\end{tabular}
\end{table}

\newpage
\textbf{Lexical selection}
\begin{table}[!h]
\centering
\renewcommand{\arraystretch}{1.3}
\begin{tabular}{l|lllll}
\toprule
\multicolumn{6}{c}{"Una \textbf{\{mask\}} situada en la región de Alta Normandía."} \\
\hline
\texttt{RoBERTa-base-BNE} &  villa &  ciudad &  localidad &  isla &  aldea \\
\texttt{RoBERTa-large-BNE} &  ciudad &  localidad &  población &  región &  villa \\
\texttt{BETO} & francesa & ciudad & localidad & población & comuna \\
\texttt{mBERT} & comuna & localidad & población & parroquia & commune \\
\texttt{BERTIN} &  región &  ciudad &  casa &  localidad &  población \\
\texttt{ELECTRA} & finca & granja & calle & ciudad & villa \\
\bottomrule
\toprule
\multicolumn{6}{c}{"Te voy a contar una \textbf{\{mask\}} sobre mi prima."} \\
\hline
\texttt{RoBERTa-base-BNE} &  historia &  anécdota &  cosa &  leyenda &  verdad \\
\texttt{RoBERTa-large-BNE} &  historia &  cosa &  anécdota &  curiosidad &  verdad \\
\texttt{BETO} & historia & cosa & pista & verdad & teoría \\
\texttt{mBERT} & novela & historia & película & pista & cinta \\
\texttt{BERTIN} &  historia &  película &  encuesta &  frase &  vez \\
\texttt{ELECTRA} & historia & película & cosa & canción & lección \\
\bottomrule
\toprule
\multicolumn{6}{c}{"Martin se \textbf{\{mask\}} para ir a pescar al río."} \\
\hline
\texttt{RoBERTa-base-BNE} &  prepara &  ofrece &  desnuda &  casa &  arregla \\
\texttt{RoBERTa-large-BNE} &  prepara &  preparaba &  levanta &  ofrece &  preparó \\
\texttt{BETO} & prepara & despierta & fue & preparó & preparan \\
\texttt{mBERT} & va & ofrece & encuentra & preparar & queda \\
\texttt{BERTIN} &  fue &  entrena &  va &  casó &  levanta \\
\texttt{ELECTRA} & usa & utiliza & prepara & usaba & emplea \\
\bottomrule
\toprule
\multicolumn{6}{c}{"Mi vida no ha sido fácil, pero yo \textbf{\{mask\}} la vida."} \\
\hline
\texttt{RoBERTa-base-BNE} &  amo &  es & , &  soy &  quiero \\
\texttt{RoBERTa-large-BNE} &  amo &  tengo &  prefiero &  vivo &  adoro \\
\texttt{BETO} & amo & soy & vivo & tengo & gano \\
\texttt{mBERT} & es & , & tiene & ama & recuerda \\
\texttt{BERTIN} &  amo &  soy &  quiero &  tengo &  gano \\
\texttt{ELECTRA} & tengo & tampoco & conozco & amo & prefiero \\
\bottomrule
\end{tabular}
\end{table}

\newpage
\textbf{Polarity agreement}
\begin{table}[!h]
\centering
\renewcommand{\arraystretch}{1.3}
\begin{tabular}{l|lllll}
\toprule
\multicolumn{6}{c}{"Llegamos muy pronto y no pude hablar con \textbf{\{mask\}}."} \\
\hline
\texttt{RoBERTa-base-BNE} &  ellos &  nadie &  vosotros &  él &  ella \\
\texttt{RoBERTa-large-BNE} &  el &  ella &  nadie &  ellos &  él \\
\texttt{BETO} & él & nadie & ella & ellos & [UNK] \\
\texttt{mBERT} & él & ellos & ella & nada & ellas \\
\texttt{BERTIN} &  D &  nadie &  ella &  S &  l \\
\texttt{ELECTRA} & nadie & él & ellos & ustedes & ella \\
\bottomrule
\toprule
\multicolumn{6}{c}{"No lo había visto \textbf{\{mask\}}."} \\
\hline
\texttt{RoBERTa-base-BNE} &  nunca &  antes &  yo &  todavía &  aún \\
\texttt{RoBERTa-large-BNE} &  nunca &  antes & . &  aún &  en \\
\texttt{BETO} & antes & nunca & así & jamás & trabajar \\
\texttt{mBERT} & él & que & ( & , & nunca \\
\texttt{BERTIN} &  él &  hoy &  ayer &  tú &  todo \\
\texttt{ELECTRA} & antes & nunca & venir & aún & todavía \\
\bottomrule
\end{tabular}
\end{table}

\section*{Appendix III}
While the main focus of the article is building language models, we also computed 300 dimensional word embeddings using FastText. Both the CBOW\footnote{\url{https://zenodo.org/record/5044988}}
and Skip-gram
\footnote{\url{https://zenodo.org/record/5046525}} 
versions are publicly available on Zenodo. Using the clean data at document level described on the previous section, the processing took around 20 days on a HPC node\footnote{\url{https://www.bsc.es/innovation-and-services/technical-information-cte-amd}}
equipped with an AMD EPYC 7742 (@ 2.250GHz) processor with 128 threads. These embeddings are not evaluated and just provided as an additional resource.

\section*{Appendix IV}
This Appendix shows several examples of the GPT2 models' text completion capabilities.

GPT2 text generation examples:
\begin{itemize}
    \item Si vas a Barcelona tines que visitar $\rightarrow$ la Sagrada Familia.
    \item En el Barcelona Supercomputing Center nos dedicamos a $\rightarrow$  todo tipo de campos que requieran de gran potencia tecnológica.
    \item El BSC junto con la BNE desarrollan un modelo del lenguaje GPT2 en español que $\rightarrow$  se podría utilizar para la gestión de la producción y la investigación científica.
    \item Sin el esfuerzo de los médicos en la pandemia no hubieramos $\rightarrow$  podido salvar la vida a los nuestros.
    \item "Yo me vacuno seguro" es la $\rightarrow$  frase que acompaña en Facebook a la imagen de Jorge.
    \item En la Universidad de Deusto $\rightarrow$  y dentro de la acción social de la universidad, se ha invitado a más de 400 personas con el fin de trabajar el voluntariado desde una perspectiva ética y humanista.
\end{itemize}

GPT2-large text generation examples:
\begin{itemize}
    \item Si vas a Barcelona tines que visitar $\rightarrow$  su iglesia, sus museos, el Modernisme (y su obra maestra el Modernismo), la estatua de Francesc de Coll, la Fuente Mágica, su teatro...
    \item En el Barcelona Supercomputing Center nos dedicamos a $\rightarrow$  impulsar y desarrollar la investigación en supercomputación.
    \item El BSC junto con la BNE desarrollan un modelo del lenguaje GPT2 en español que $\rightarrow$  permitirá estudiar el lenguaje desde un enfoque de lenguaje natural.
    \item Sin el esfuerzo de los médicos en la pandemia no hubieramos $\rightarrow$  podido salvar a los enfermos.
    \item "Yo me vacuno seguro" es la $\rightarrow$  frase que ha escogido un joven de 24 años.
    \item En la Universidad de Deusto $\rightarrow$  nos gusta pensar que tenemos que estar muy al día en todo para poder adaptarnos al ritmo de los tiempos en los que vivimos.
\end{itemize}

\end{document}